\documentclass[11pt,a4paper]{article}
\usepackage[hyperref]{tex/acl2018}
\usepackage{times}
\usepackage{latexsym}
\usepackage{subcaption}
\usepackage{graphicx}
\usepackage[colorinlistoftodos]{todonotes}
\usepackage{url}

\def\sem#1{[\![ \; #1 \; ]\!]}

\title{Specificity measures and reference}

\author{Albert Gatt \\
  Inst. of Linguistics \& Language Technology \\
  University of Malta \\
  {\tt albert.gatt@um.edu.mt} 
  \\\And
  Nicol\'{a}s Mar\'{i}n \\
  Dept. of Computer Science \& AI \\
  University of Granada\\
  {\tt nicm@decsai.ugr.es} 
  \\ \AND
  Gustavo Rivas-Gervilla \\
  Dept. of Computer Science \& AI \\
  University of Granada\\
  {\tt griger@decsai.ugr.es} 
  \\ \And
  Daniel S\'{a}nchez \\
  Dept. of Computer Science \& AI \\
  University of Granada\\
  {\tt daniel@decsai.ugr.es}}

\date{}
\aclfinalcopy
\begin{document}
\maketitle

\begin{abstract}
In this paper we study empirically the validity of measures of referential success for referring expressions involving gradual properties.
More specifically, we study the ability of several measures of referential success to predict the success of a user in choosing the right object, given a referring expression. Experimental results indicate that
certain fuzzy measures of success are able to predict human accuracy in reference resolution.
Such measures are therefore suitable for the estimation of the success or otherwise of a referring expression produced by a generation algorithm, especially in case the properties in a domain cannot be assumed to have crisp denotations.
\end{abstract}

\section{Introduction}\label{sec:intro}
Referring expression generation ({\sc reg}) is one of the subtasks of Natural Language Generation ({\sc nlg}) systems. Given a context comprised of a set of objects and a collection of properties that can be predicated of those objects, the {\sc reg} problem is to find referring expressions -- that is, subsets of those properties -- that allow a user to locate a specific object and distinguish it from its distractors \cite{Dale1995,Krahmer2012,vanDeemter:2016}. 

The task definition given above corresponds to the {\em content determination} part of {\sc reg}; in standard accounts, {\sc reg} also involves a realisation step in which the form of a referring expression needs to be determined \cite{CastroFerreira2018}. 

From a general point of view, the {\sc reg} problem intends to emulate through automatic means the process carried out by a human being whose purpose is to identify a certain object in a specific context, using natural language, so that another receiving user is able to precisely and univocally identify it. 

One source of complexity for {\sc reg} is context-dependence and graduality. This has become increasingly evident in work that has sought solutions to the {\sc reg} problem in naturalistic scenes, as part of a broader research focus on the vision-language interface \cite{Kazemzadeh2014,Mao2016,Yu2016}. However, context dependence is also a central concern for approaches to {\sc reg} that assume a more structured input representation where entities and their properties are available, but the extent to which a property applies to a referent is not necessarily an all-or-none decision \cite{Horacek2005,VanDeemter2006,Turner2008a,Williams2017}. Under these conditions, it is no longer possible to assume that properties are crisp or Boolean, or even that both sender and receiver necessarily assume the same semantics for those properties. For example, it may not be realistic to assume that all objects are red to the same degree, or that both sender and receiver have the same model of what counts as `red'. Furthermore, the utility of the term `red' in identifying the referent will depend in part on context, that is, whether there are any other red entities, and whether they are red to the same extent. 

The above issues affect how referential success is to be defined, that is, what criteria an algorithm should use to determine whether a candidate referring expression is likely to succeed in helping a receiver identify the intended referent. There are three questions that arise in this connection: 1) How to model the semantics of gradual properties, 2) how to compute the degree of success of a candidate referring expression based on those semantics; and 3) whether such a measure of the degree of success is valid, in the sense that it indeed correlates with the ease and success with which a receiver in fact resolves the reference.


In this work we present the results of an empirical study of these issues,
using geometric objects and basic properties such as color, size and position, all of which can be viewed as gradual. As we shall see, in relation to question number 1, we take results from the theory of fuzzy sets for the modeling of properties. In relation to question 2, the use of eight gradual measures related to referential success, which derive from the well-known concept of specificity of possibility distributions, is proposed. In relation to the third, an experimental study is made to validate these measures, as well as to gain a better understanding of the factors influencing referential success in the presence of gradual properties. The results are encouraging, in that they show that specificity measures do predict behavioural outcomes and hence that the framework proposed here provides a plausible account of success that can be incorporated in existing {\sc reg} algorithms.

The rest of this paper is organized as follows: Section 2 introduces the problem of referential success of referring expressions with gradual properties, and the role specificity measures can play in this setting. In the same section we describe the set of measures that will be employed in the study. Section 3 describes an experiment to evaluate whether such specificity measures are useful in predicting the success of a referring expression. The results of the experiment are presented in Section 4. Finally, Section 5 is devoted to some discussion and conclusions.

\section{Background}\label{sec:measures}
A referring expression $re = \left\{p_{1}, \ldots, p_{n}\right\}$ is said to have referential success for a certain object $o$ when the object is the only one satisfying all the properties in the expression $re$. Formally:

\begin{equation}
\label{eq:ref-succ-crisp}
\bigcap_{p \in re} \sem{p} = \left\{o\right\}
\end{equation}
where $\sem{p}$ is the set of objects satisfying $p$.

When the properties are Boolean the above expression is both Boolean and easy to compute. Many {\sc reg} algorithms are in fact couched as search procedures, where the process of searching for a combination of properties terminates as soon as the criterion in (\ref{eq:ref-succ-crisp}) is fulfilled \cite{Dale1995,Krahmer2012}. Additionally, since it is assumed that both sender and receiver have identical interpretations of the semantics of the relevant properties, 
it is accepted that when Eq. (\ref{eq:ref-succ-crisp}) holds, the receiver should be able to identify the target object, at least in principle. In practice, the ease and/or speed with which references are both produced and resolved by humans depends also on whether the choice of properties conforms to preferences, salient characteristics of objects, etc \cite{Pechmann1989,Tarenskeen2015,Rubio-Fernandez2016}.

However, in many application domains we work with gradual properties instead of Boolean ones \cite{ipmu,VanDeemter2006,Turner2008a}. This means that the success criterion in \ref{eq:ref-succ-crisp}) no longer applies, unless the properties in question are transformed into crisp ones, as is done for example in the conversion of numerical properties to inequalities by \citet{VanDeemter2006}. This, however, is not easily applicable to properties that are not typically modelled as numeric, but which are still gradual. Examples include colour (to what extent is the object red?) or location (how far towards the top or the left is the object?).
 
\subsection{Properties as fuzzy sets}
In this paper, the theoretical starting point for a treatment of referential success as gradual is a set of insights developed within the field of Fuzzy Set Theory and used to determine the core properties that measures of referential success need to satisfy \cite{MarinEtAl2016:WCCI}.

In Fuzzy Set Theory, gradual properties (i.e. properties in which objects have varying degrees of membership) are modelled using possibility distributions and associated with linguistic terms through tools such as linguistic variables or, more broadly, the results of the Computational Theory of Perceptions~\cite{TCP}. 

Possibility distributions are fuzzy sets that represent the available information about the actual (unique) value of a given variable. In the context that concerns us, these types of functions can be used to represent the available information about what object a given expression refers to, whereby degrees of possibility indicate that some values are more plausible than others. 

Thus, in order to analyze the referential success of the expression, it is useful to determine how difficult it is to find out the actual value (i.e. the object referred to) among those that belong to the associated possibility distribution to a greater or lesser degree. 
As a particular case, crisp sets can be employed for such purpose, all values being equally and completely plausible. For instance, one may say that the value of a variable $X$ is in $\{3,5,6\}$, but we don't know which value in the set is the actual value of $X$.

\subsection{The concept of specificity}

For general (fuzzy) possibility distributions, the well-known measures of  specificity~\cite{Yager1982-specificity} allow to determine how easy it is to determine the real value in view of the possibility distribution (that is, to which extent the distribution \emph{specifies} the value). This can be interpreted in some contexts as the extent to which a possibility distribution is a singleton (i.e. the possibility distribution clearly indicates a unique value). There are infinitely many specificity measures, that can be classified into different families (linear, product, etc.). They serve to assess the amount of uncertainty about the value of the variable. As an example in the crisp case, the set $\{3,5,6\}$ is less specific than the set $\{3,5\}$, and hence $X \in \{3,5,6\}$ is a more uncertain statement than $X \in \{3,5\}$. Several additional theoretical results have been adduced based on this core insight ~\cite{Dubois-Prade1987-specificity,Yager1990-specificity,Yager1992-specificity,Garmendia2003,MarinRSY17}.

The definition of specificity measures suggests a clear connection with
the concept of referential success that is particularly useful when fuzzy properties are involved: A referring expression $re$ has referential success to the extent that 1) the (fuzzy) set of objects satisfying $re$, denoted by $O_{re}$, viewed as a possibility distribution, is `specific' (in the sense explained above), and ii) $O_{re}$ contains the object intended to be referred. 

Hence, the referential success of an expression $re$ is upper-bounded by the specificity of the associated fuzzy set and the fulfillment of $re$ by the intended object.
In recent theoretical work, it has been shown that specificity measures can be used to derive measures of referential success for expressions containing gradual properties~\cite{MarinEtAl2016:WCCI}, and conversely, that referential success measures can derive specificity measures~\cite{IEEETRAN-Marin-RivasGervilla-Sanchez-Yager}.

\subsection{Spsecificity measures}

\begin{table*}[!h]
  \begin{subtable}[t]{0.45\textwidth}
  \small
      \centering
         \begin{tabular}{|c|c|}
              \hline
              \textbf{Name}&Definition\\
              \hline
              $m_1$&$a_1-a_2$\\
              $m_3$&$a_1*(a_1-a_2)$\\
              $m_5$&$a_1*(1-a_2)$\\
              $m_7$&$min\{1,1-a2\}$\\
              \hline
          \end{tabular}
          \caption{Specificity measures}
          \label{tabmeasures}
  \end{subtable}    
  ~
\begin{subtable}[t]{0.45\textwidth}
	\centering
    \small
		\begin{tabular}{|c|c|}
			\hline
			\textbf{Name}&Definition\\
		\hline
		$m_2$&$(a_1-a_2)/a_1$\\
		$m_4$&$a_1*(a_1-a_2)/a_1$\\
		$m_6$&$a_1*(1-a_2)/a_1$\\
		$m_8$&$min\{1,1-a2\}/a_1$\\	
		\hline
	\end{tabular}
    \caption{Normalised versions of measures in Table \ref{tabmeasures}.}
	\label{tabmeasures2}
	\end{subtable}
    \caption{Specificity measures and their normalised counterparts}
    \label{table:measures}
\end{table*}


In view of the previous discussion, as indicated by \citet{MarinRS17}, given a fuzzy set associated with a referring expression and a set of objects in a given domain of discourse, the specificity of the fuzzy set indicates to what extent there is some single object of which the expression is true,
that is, to what extent this expression has referential success in the set for some (unknown) object. Since there is an infinity of possible specificity measures, an important question is which family of measures is useful and empirically valid in the context of {\sc reg}.
We shall restrict our experimental study to the validity of a few measures that have been shown to be the most suitable for our purposes from a theoretical perspective \cite{MarinRSY17}.

Let $\mathcal{O}$ be the universe of objects in a given context. Let $A \in [0,1]^\mathcal{O}$ with $|\mathcal{O}|=m$ be the possibility distribution associated with a given referring expression. Let us consider that memberships of objects in $A$ are ranked as $a_1 \geq a_2 \geq \cdots \geq a_m$.

Taking into account the previous framework, the specificity measures considered in this work are shown in table \ref{tabmeasures}. All these measures satisfy an important property for our purposes: when there are two objects that comply with the referring expression to degree 1 (that is, when $a_2=1$), the measures yield 0, which is a desirable result in terms of referential success (since it indicates that the expression is applicable to at least two objects, rather than just the target). Intuitively, this implies that if a target referent with property $p$ has a similar distractor with $p$, a referring expression that uses $p$ will be less successful.

The value computed by these measures can be thought of as reflecting the distance between $a_2$ and $a_1$, that is the distance, in terms of their membership in $re$, between the second-ranked and the highest-ranked entity in $re$ of the second-ranked entity $a_2$, from the 
This is due to the fact that in all the cases, specificity is upper bounded by the value of $a_1$. As a consequence, in the particular case that the fuzzy set $A$ is not normalized (that is, when $a_1<1$), the measure can yield the same value for different situations.  For instance, $m_1$ in Table \ref{table:measures}, yields the same value in case $a_1=1$, $a_2=0.5$ and in case $a_1=0.5$, $a_2=0$. 

In order to distinguish these two cases, we propose to use measures that work as indices of the relation between the measures in Table \ref{table:measures}, and the value of $a_1$. These are shown in table \ref{tabmeasures2}, and are defined as follows: $m_i$ in Table \ref{tabmeasures2} is obtained from $m_{i-1}$ in Table \ref{tabmeasures} as $m_{i-1}/a_1$, assuming by convention that $a_1=0$ implies $m_i=0$ for all measures in Table \ref{tabmeasures2}. All measures in Table \ref{tabmeasures2} are in $[0,1]$. Note that in case $a_1=1$, that is, when there is at least one object that fully complies with the referring expression, measures $m_i$ in Table \ref{tabmeasures2} coincide with the corresponding measures $m_{i-1}$ in Table \ref{tabmeasures}. As a final remark, note that $m_1=m_4$.


\begin{figure*}
\centering
    \begin{subfigure}[b]{0.45\textwidth}
        \includegraphics[width=\textwidth]{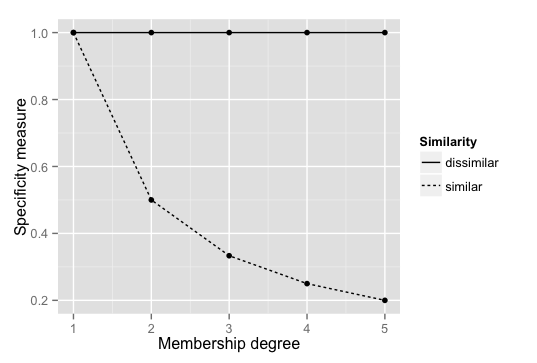}
        \caption{Measure $m_{2}$}
        \label{fig:m2}
    \end{subfigure}
    ~
    \begin{subfigure}[b]{0.45\textwidth}
        \includegraphics[width=\textwidth]{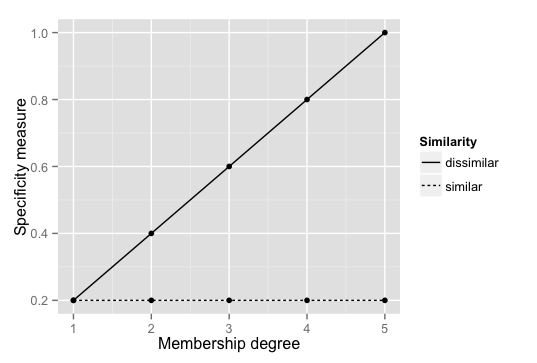}
        \caption{Measure $m_{4}$}
        \label{fig:m4}
    \end{subfigure}
    
    \begin{subfigure}[b]{0.45\textwidth}
        \includegraphics[width=\textwidth]{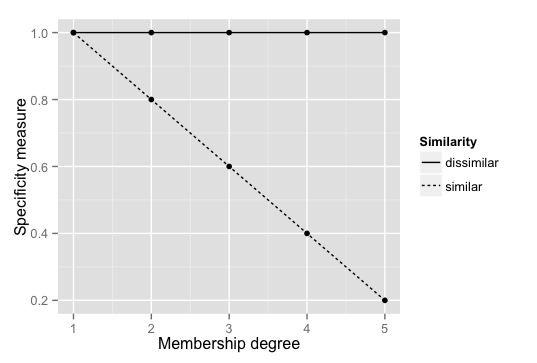}
        \caption{Measure $m_{6}$}
        \label{fig:m6}
    \end{subfigure}
    ~
    \begin{subfigure}[b]{0.45\textwidth}
        \includegraphics[width=\textwidth]{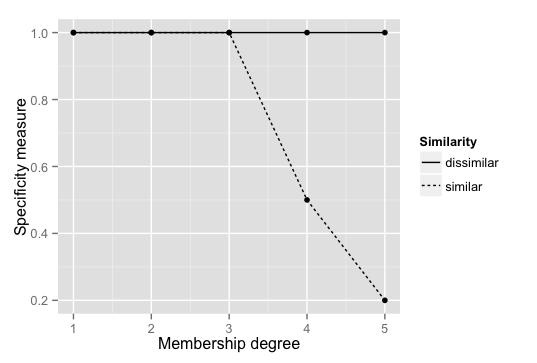}
        \caption{Measure $m_{8}$}
        \label{fig:m8}
    \end{subfigure}
\caption{Specificity measures for the property Colour, by similarity and membership degree}
\label{fig:spec-ex}
\end{figure*}

\section{Validating the specificity measures: An experiment}\label{sec:experiment}
We conducted an empirical study with a view to addressing the third question highlighted in the introduction, that is, to assess the validity of the specificity measures introduced in the previous section. The study took the form of an experiment in which the task was to identify a referent in a visual domain given a referring expression. This yielded two behavioural measures, identification time (id-time) and accuracy, both of which have been previously used in task-based {\sc reg} evaluations \cite{Gatt2010}. Our aims were twofold: on the one hand, the experiment was intended as an empirical investigation of the impact of certain variables, notably gradability of properties, on the success of referring expressions; on the other, the data serves as a testbed to see whether the variance in id-time and (probability of) accuracy could be predicted by the measures in Table \ref{table:measures}.

\subsection{Participants}
Twenty-one participants (17 male, age range 21--54; median age 26), with different academic profiles (mostly staff of the University of Granada), 
took part in the study. Participation was voluntary.

\subsection{Materials and design}
Each experimental item consisted of a visual display consisting of 5 geometric shapes (triangles, circles and/or squares), one of which was the target referent, which was accompanied by a referring expression which mentioned one property in addition to the head noun. There were four possible properties, for each of which 5 degrees of membership distinct from 0 were defined:

\begin{enumerate}
\itemsep0em 
\item Colour: Using the HSB colour space, all colours had constant hue and brightness, with membership defined in terms of saturation.
\item Size: Three different size intervals (small, medium and large) were defined. An object's maximum membership in a given size was defined in terms of whether the object fell towards lower (resp. higher) extremes of the interval in the case of small (resp. large), or towards the middle of the interval in the case of medium size.
\item Vertical location: The screen height was divided into three equal-sized intervals, corresponding to bottom, middle and top. Once again, membership was highest for objects close to the extremes of the intervals in the case of bottom and top, and towards the middle of the interval otherwise.
\item Horizontal location: Treated as above, by dividing the width of the screen into three equal intervals.
\end{enumerate}

A further factor was {\em similarity}, which referred to whether there was one distractor in the display which was similar to the target on the identifying property. Let $m$ be the degree of membership of some referent $o$ in some identifying property $p$, and let $d$ be one of the distractors in the visual domain, with membership $m'$ in $p$. Then, $o$ and $d$ were defined as {\em similar} if $\vert m - m' \vert \leq \alpha$ for some threshold $\alpha$, and {\em dissimilar} otherwise. Thresholds were set differently for the four properties. 

The above is a $4$ (property) $\times$ $2$ (similarity) $\times$ $5$ (membership) within-subjects design. We created 40 different target items with different shapes and colour/size/vertical/horizontal combinations. Items were rotated through a latin square so that different participants saw different items in different conditions. 

\subsection{Relationship to specificity measures}
For a given experimental item involving a referring expression containing property $p$, it is possible to compute the specificity of the expression based on the degree of membership of the target referent in $p$. As an example, Figure \ref{fig:spec-ex} plots the normalised measures in Table \ref{tabmeasures2} against membership degree for the two different values of similarity, for the property colour (the plots are in fact identical for all properties). Note that in the dissimilar case, where no distractors are present that are close to the referent, the specificity is either maximal or linearly increasing with membership. The opposite is true for the similar condition.

\begin{figure}[!t]
\centering
\includegraphics[scale=0.13]{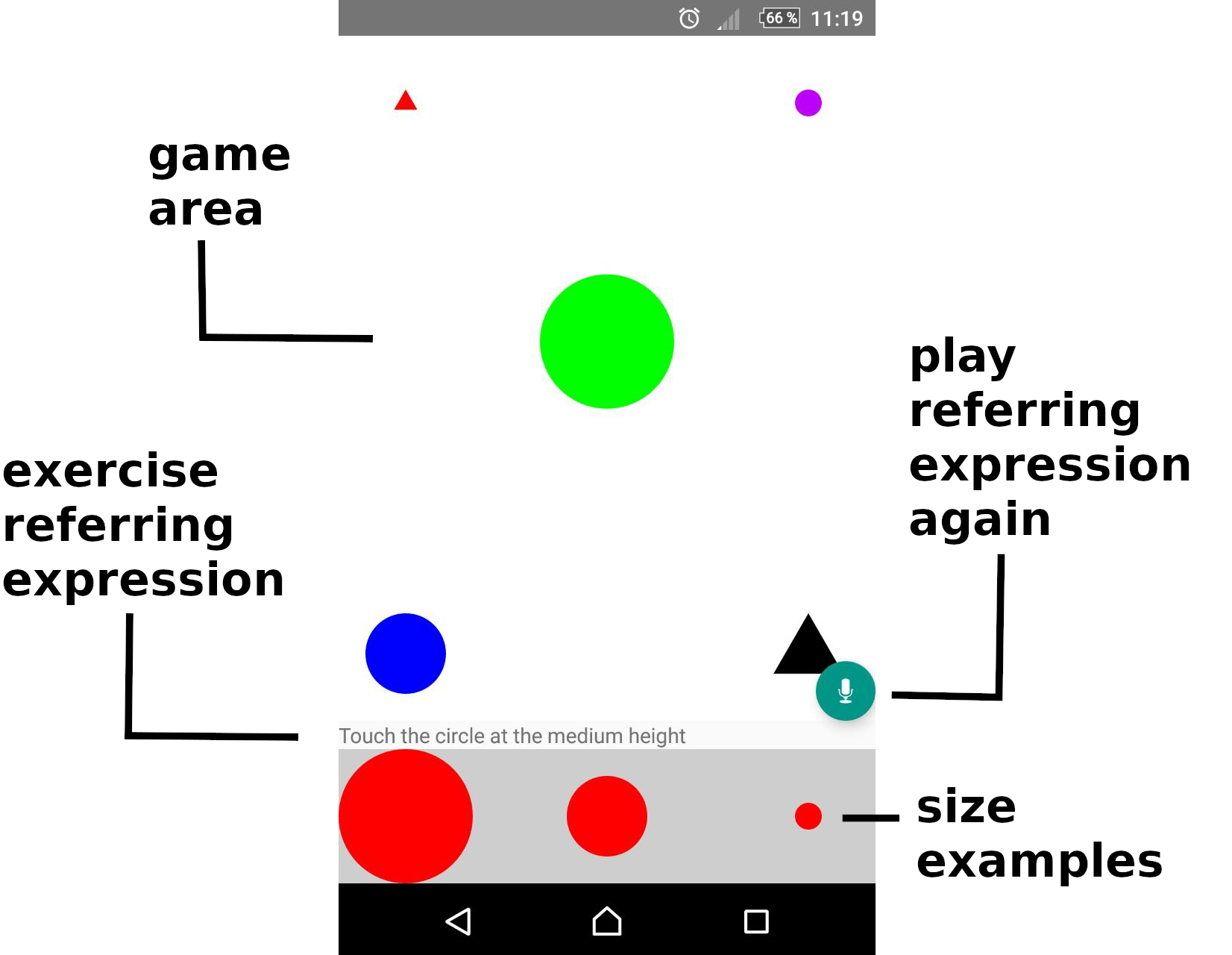}
\caption{Screenshot of the experiment app. The referring expression in this case is {\em the circle at the medium height}.}
\label{fig:screenshot}
\end{figure}

\subsection{Procedure}
The experiment was implemented as part of an Android application called Refer4Learning \cite{MarinRS17}, originally designed to help teachers in the early stages of child education to work on basic concepts such as color, size, or position of simple geometric objects. For the purposes of the present experiment, a version of the app was created for use with adult participants to administer experimental trials.

As shown in Figure \ref{fig:screenshot}, trials consisted of a visual display 
with an instruction of the form {\em touch the NP}, where {\em NP} described the intended referent, which contained the identifying property (e.g. {\em the circle at the \underline{middle height}}). The instruction was also played as an audio file.

Of the four properties used, size required special treatment. While the degree of membership in a colour or location property can usually be determined based on the visual configuration or the user's knowledge (e.g. an object in a shade of pink has less membership in red than a stereotypically red object), for size, one typically requires reference points to determine what counts as stereotypically large, small etc, or else there is complete reliance on a comparison to other objects in the domain \cite{VanDeemter2006}.

In the present case, this would result in a potential confusion between membership degree (to what extent is this object large?) and similarity (how much larger is it than others?). Hence, a standard of comparison was provided for expressions using size, showing three representative sizes for the target shape. An example is shown at the bottom of Figure \ref{fig:screenshot}, involving circles (since the target referent and distractors happen to be circles in this instance).

The app recorded the object a participant identified as well as the time taken to identify it. Each participant saw the 40 trials in random order.  
\begin{figure*}[!t]
\centering
    \begin{subfigure}[b]{0.45\textwidth}
        \includegraphics[scale=0.35]{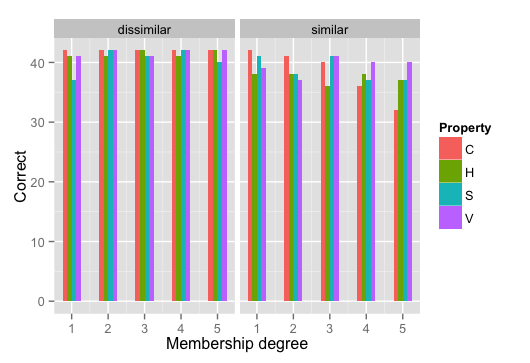}
        \caption{Accuracy}
        \label{fig:acc-prop}
    \end{subfigure}
    ~ 
    \begin{subfigure}[b]{0.45\textwidth}
        \includegraphics[scale=0.35]{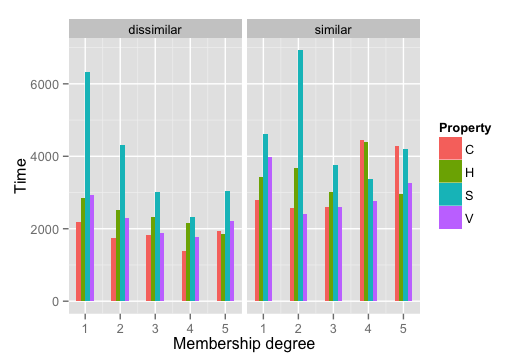}
        \caption{Identification time (id-time)}
        \label{fig:time-prop}
    \end{subfigure}
\caption{Time and accuracy as a function of property, similarity and membership (best viewed in colour).}
\label{fig:acc-time-prop}
\end{figure*}

\begin{figure*}[!t]
\centering
    \begin{subfigure}[b]{0.45\textwidth}
        \includegraphics[scale=0.35]{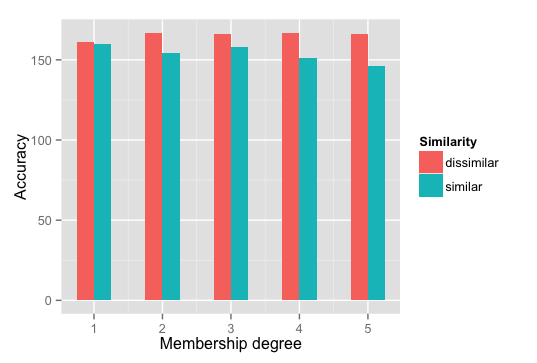}
        \caption{Accuracy}
        \label{fig:acc}
    \end{subfigure}
    ~ 
    \begin{subfigure}[b]{0.45\textwidth}
        \includegraphics[scale=0.35]{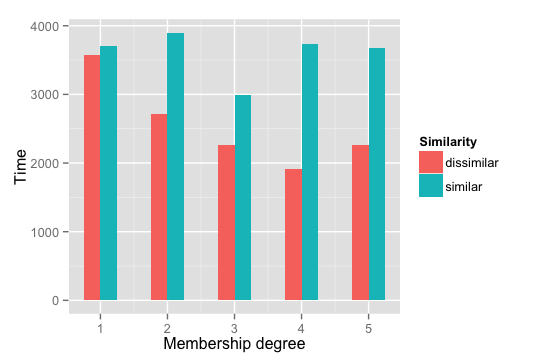}
        \caption{Identification time (id-time)}
        \label{fig:time}
    \end{subfigure}
\caption{Time and accuracy as a function of similarity and membership (best viewed in colour).}
\label{fig:acc-time}
\end{figure*}

\section{Results}\label{sec:results}
The analysis proceeds in two stages. First, we investigate the impact of the fixed effects of property, similarity and membership on participant responses as reflected in their id-time and accuracy. Second, we analyse the relationship between the primary measures of specificity and indexes presented in Section \ref{sec:measures} and the results.

\subsection{Effects of membership degree, similarity and property}
Figure \ref{fig:acc-prop} summarises the frequency of correct responses, according to the identifying property, the similarity condition, and the target's degree of membership in the identifying property. The same is shown for mean reaction time in Figure \ref{fig:time-prop}.

These figures suggest that similarity and membership degree had an impact on both accuracy and id-time. However, the picture varies considerably from one property to another. For example, accuracy goes down with increasing membership when the identifying property is colour and there is a similar distractor, while id-time is faster for size as membership increases, particularly in the dissimilar condition.

Figure \ref{fig:acc-time} shows the impact of membership and similarity more clearly, by aggregating over all the four levels of the property fixed effect. Figure \ref{fig:acc} shows that accuracy of identification is lower when there is a similar distractor to the target. In the case of id-time (Figure \ref{fig:time}), participants took longer to identify the target in the presence of a similar distractor. While membership degree does not appear to exert a clear impact on time in the similar case, in the dissimilar case, there is a downward trend in id-time as membership degree increases. This suggests that, in the absence of a similar distractor, people resolved references faster the more clearly the target belonged to the identifying property.

In the remainder of this part of the analysis, we use hierarchical mixed models, conducting an analysis separately for accuracy (modelled as a binomial variable) and time. In each case, we first build a separate model with each of the three factors as fixed effect, and establish whether the fixed effect helps to predict the dependent by comparing this model's goodness-of-fit to that of a null model consisting of only an intercept. We then construct a full model combining all fixed effects and their interactions and report the outcomes of a significance test. Models for accuracy are logit mixed models; those for id-time are linear mixed models. All models include by-subject and by-item random intercepts. Models are compared using log likelihood and the Bayesian Information Criterion ({\sc bic}; lower values indicate better fit). Membership is modelled as a continuous predictor; similarity and property are centered around a mean of zero to facilitate interpretation of main effects. Models are built using the {\tt lme4} package in R \cite{Bates2015}; significance testing is conducted with {\tt lmerTest} \cite{Kuznetsova2014}.

\begin{table}
	\centering
  \begin{subtable}[t]{0.45\textwidth}
  	\small
	\begin{tabular}{|l|ccc|}
		\hline
		Model	&	{\sc bic}	&	{\sc ll}	&	$\chi^{2}$\\
		\hline
        0. Intercept	&	546.74	&	-263.27	&			\\
        1. Similarity	&	502.31	&	-237.69	&	51.159$^{***}$		\\
        2. Membership	&	550.34	&	-261.7	&	3.134${.}$	\\
        3. Property	&	552.88	&	-262.97	&	0.5939	\\
        \hline
	\end{tabular}    
    \caption{\small Accuracy}
    \label{table:simple-models-acc}
  \end{subtable}
  ~
    \begin{subtable}[t]{0.45\textwidth}
    	\small
    	\begin{tabular}{|l|ccc|}
		\hline
		Model	&	{\sc bic}	&	{\sc ll}	&	$\chi^{2}$\\
		\hline
        0. Intercept	&	16019	&	-7996.1	&			\\
        1. Similarity	&	16003	&	-7984.5	&	23.313$^{***}$\\
        2. Membership	&	16020	&	-7993.4	&	5.4181$^{*}$\\
        3. Property	&	16024	&	-7995	&	2.2904\\
        \hline
        \end{tabular}
    \caption{\small Identification time}
    \label{table:simple-models-time}
  \end{subtable}
	\caption{Single-factor model comparisons. All comparisons are to Model 0, the null model with no fixed effects. {\sc bic}: Bayesian Information Criterion; {\sc ll}: Log Likelihood. $^{***}$: significant at $p < .001$; $^{*}$: significant at $p < .05$; $^{.}$: marginally significant at $p \geq .05$.}
    \label{table:simple-models}
\end{table}

Table \ref{table:simple-models} shows the simple model comparisons for both accuracy and time. These suggest that of all the single predictors, it is similarity that has the highest explanatory value. The likelihood of participants identifying the correct target can to some extent be predicted from the membership degree of the target in the identifying property, though this model fits the data only marginally better than the null model. The role of membership degree is more clear in the case of id-time. There is no obvious impact of the type of property used to identify the target referent. 

\begin{table*}
\centering
\begin{subtable}[t]{0.45\textwidth}
\small
\begin{tabular}{|l|ccc|}
\hline
Fixed Effects	&	Estimate	&	Std. Error	&	$z$-value		\\
\hline
(Intercept)	&	2.88	&	0.45	&	6.39$^{***}$	\\
S	&	0.07	&	0.74	&	0.09		\\
M	&	0.07	&	0.13	&	0.53		\\
P	&	-0.66	&	0.33	&	-2.01$^{*}$	\\
S $\times$ M	&	-0.73	&	0.26	&	-2.77$^{**}$	\\
S $\times$ P	&	-0.42	&	0.66	&	-0.64		\\
M $\times$ P	&	0.18	&	0.12	&	1.58		\\
S $\times$ M $\times$ P	&	0.29	&	0.23	&	1.26		\\
\hline
\end{tabular}
\caption{\small Accuracy }
\label{table:full-acc}
\end{subtable}
~
\begin{subtable}[t]{0.45\textwidth}
\small
\begin{tabular}{|l|ccc|}
\hline
Fixed Effects	&	Estimate	&	Std. Error	&	$t$-value		\\
\hline
(Intercept)	&	3620.53	&	329.01	&	11$^{***}$	\\
S	&	119.29	&	508.24	&	0.24		\\
M	&	-182.07	&	76.62	&	-2.38$^{*}$	\\
P	&	585.02	&	232.54	&	2.52$^{*}$	\\
S $\times$ M	&	314.26	&	153.24	&	2.05$^{*}$	\\
S $\times$ P	&	102.32	&	454.59	&	0.23		\\
M $\times$ P	&	-142.17	&	69.58	&	-2.04$^{*}$	\\
S $\times$ M $\times$ P	&	-125.16	&	137.06	&	-0.91\\
\hline
\end{tabular}
\caption{\small Identification time (id-time)}
\label{table:full-time}
\end{subtable}
\caption{Full models incorporating all fixed effects and interactions. Legend: $^{***}$ significant at $p < .001$; $^{**}$ significant at $p < .01$; $^{*}$ significant at $p < .05$.}
\label{table:full-models}
\end{table*}

Next, we combine all fixed effects and their interactions in a single model to predict accuracy and time. The breakdown of the models is shown in Table \ref{table:full-models}. In both models, the effect of similarity turns out to be due to its interaction with membership degree. This corresponds to observations made above in connection with Figure \ref{fig:acc-time}: As membership degree increases, participants were more likely to identify the wrong target, and were slower in responding, in case there was a similar distractor. 

On the other hand, a main effect of property is also evident. The nature of this main effect can be interpreted with reference to Figure \ref{fig:acc-time-prop} where, as noted above, the impact of membership degree and similarity differs between properties. In the case of id-time, membership interacts significantly with property type. This is likely due to the tendency for participants to respond faster as membership increases, when size or horizontal location is the identifying property (cf. Figure \ref{fig:time-prop}).

\subsection{Validation of specificity measures}
In order to validate the specificity measures, we address two questions: 1) to what extent does the specificity of a referring expression, computed using one of the measures in Table \ref{table:measures}, predict the accuracy and speed with which participants resolve references?; 2) do the specificity measures correlate negatively with id-time, as would be expected?

\begin{table}[!h]
\begin{subtable}[b]{0.45\textwidth}
\centering
\small
\begin{tabular}{|l|ccc|}
\hline
	&	{\sc bic}	&	{\sc ll}	&	$z-$test		\\
\hline    
m1	&	508.54	&	-240.81	&	4.871$^{**}$	\\
m2	&	500.68	&	-236.87	&	6.92$^{**}$	\\
m3	&	532.98	&	-253.02	&	3.809$^{**}$	\\
m4	&	508.54	&	-240.81	&	4.871$^{**}$	\\
m5	&	517.13	&	-245.1	&	4.934$^{**}$	\\
m6	&	504.32	&	-238.69	&	6.851$^{**}$	\\
m7	&	528.52	&	-250.79	&	4.621$^{**}$	\\
m8	&	516.73	&	-244.9	&	6.191$^{**}$	\\
\hline
\end{tabular}
\caption{Dependent: Accuracy (binomial)}
\end{subtable}
~
\begin{subtable}[b]{0.45\textwidth}
\centering
\small
\begin{tabular}{|l|ccc|}
\hline
	&	{\sc bic}	&	{\sc ll}	&	$t-$test		\\
\hline    
m1	&	15994	&	-7980.4	&	-5.662$^{**}$	\\
m2	&	16013	&	-7989.8	&	-3.581$^{**}$	\\
m3	&	16002	&	-7984.4	&	-4.883$^{**}$	\\
m4	&	15994	&	-7980.4	&	-5.662$^{**}$	\\
m5	&	15996	&	-7981	&	-5.555$^{**}$	\\
m6	&	16015	&	-7990.9	&	-3.25$^{*}$ 	\\
m7	&	15998	&	-7982.4	&	-5.277$^{**}$	\\
m8	&	16018	&	-7992.4	&	-2.738$^{*}$	\\
\hline
\end{tabular}
\caption{Dependent: Response time}
\end{subtable}
\caption{Models with each of the eight success measures as predictor. {\sc bic}: Bayesian Information Criterion; {\sc ll}: Log-likelihood. $^{**} p < .01$; $^{*} p < .01$.}
\label{table:measures-models}
\end{table}

In order to check whether the specificity measures have predictive power, we conducted two sets of mixed-models analysis, one on accuracy and one on id-time. In each case, constructed models contain one of the measures as the sole predictor. The outcomes are summarised in Table \ref{table:measures-models}. All measures emerge as significant predictors of the likelihood with which a participant identifies a target referent accurately, and of the variance in id-time. 

\begin{table}[!h]
\small
\centering
\begin{tabular}{|l|ccc|}
\hline
  & $\bar{X}_{corr}$ & $\bar{X}_{incorr}$ & $\bar{X}_{corr} - \bar{X}_{incorr}$\\
  \hline
m1	&	0.418	&	0.238	&	0.180	\\
m2	&	0.756	&	0.476	&	0.280	\\
m3	&	0.293	&	0.161	&	0.132	\\
m4	&	0.418	&	0.238	&	0.180	\\
m5	&	0.455	&	0.301	&	0.154	\\
m6	&	0.824	&	0.588	&	0.235	\\
m7	&	0.493	&	0.360	&	0.134	\\
m8	&	0.890	&	0.689	&	0.201	\\
\hline
\end{tabular}
\caption{Accuracy: Means and differences for the 8 specificity measures. $\bar{X}_{in/corr}$: mean measure of specificity for in/correctly identified referents.}
\label{table:acc-means}
\end{table}

\begin{table}[!h]
\centering
\small
\begin{tabular}[t]{|l|cc|}
\hline
& Pearson's $r$ & Spearman's $\rho$\\
\hline
m1	&	-0.184	&	-0.258	\\
m2	&	-0.117	&	-0.202	\\
m3	&	-0.159	&	-0.239	\\
m4	&	-0.184	&	-0.258	\\
m5	&	-0.181	&	-0.293	\\
m6	&	-0.107	&	-0.202	\\
m7	&	-0.172	&	-0.265	\\
m8	&	-0.091	&	-0.163	\\
\hline
\end{tabular}
\caption{Correlation coefficients between time and each referential success measure}
 \label{table:time-corr}
\end{table}

\paragraph{Accuracy} To investigate the role of different measures in predicting accuracy, Table \ref{table:acc-means} divides the experimental data into instances on which a participant correctly identified the target referent and those where they did not. For each, we consider the mean of each specificity measure and the difference in means. The latter can be thought of as a measure of `sensitivity': The greater the difference between means of correct vs. incorrect trials, the more a measure is able to differentiate between the two. On these grounds, we obtain an ordering of the measures as follows: m3 \textless{} m7 \textless{} m5 \textless{} m1 \textless{} m4 \textless{} m8 \textless{} m6 \textless{} m2.

\paragraph{Identification time} We further investigated the correlation between id-time and each specificity measure using both Pearson and Spearman correlations. These are summarised in Table \ref{table:time-corr}. The correlations go in the predicted direction (i.e. they are negative), though they are generally on the low side. The low coefficients suggest that the relationship between time and specificity is non-linear; in the case of Pearson's $r$, a further assumption that is probably violated in our data is that of monotonicity. 

\begin{figure}[!h]
\centering
\includegraphics[scale=0.5]{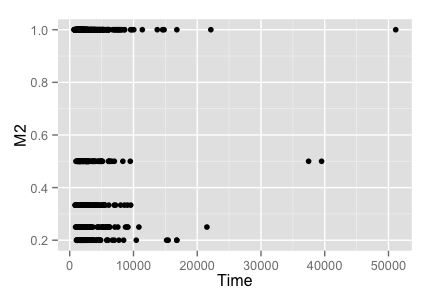}
\caption{Relationship between $m2$ and id-time}
\label{fig:corr-time-plot}
\end{figure}

As one example, Figure \ref{fig:corr-time-plot} plots id-time against measure $m2$ (similar patterns obtain for all measures). What the figure suggests is a tendency for the measure to divide cases into classes which do not linearly correspond with time. This is in spite of the evidence in Table \ref{table:measures-models}, that specficity measures are good predictors of the variance in reaction time, as well as accuracy.

Ranking the measures according to the Spearman correlation coefficient yields a different ordering from the one obtained for accuracy: m5 \textless{} m7
\textless{} m1 \textless{} m4 \textless{} m3 \textless{} m2 \textless{}
m6 \textless{} m8.

In summary, the results indicate that referential success measures can reliably predict human performance in resolving referring expressions. Crucially, however, the relationship is strongest with respect to accuracy, rather than id-time. This is to be expected: our measures of referential success are after all intended as measures of how likely it is that a referring expression singles out a particular object.

\section{Discussion and Conclusions}\label{sec:discussion}
This paper addressed the notion of referential success, arguing that predicting the success of a referring expression generated by a {\sc reg} algorithm needs to take into account the graduality of the available properties and contextual factors, including the degree of membership in a property of a target's distractors.

While gradable properties have been addressed in the {\sc reg} literature \cite{VanDeemter2006}, the implications of graduality for referential success remain under-explored and the complex interactions between the evaluation of the properties of an entity and the visual context it is in have only recently begun to benefit from principled accounts \cite{Yu2016,Williams2017}.

The account of referential success given here is based on the concept of specificity, a measure of the identifiability of an entity based on the possibility distribution associated with its properties. Hence, we do not consider any property as necessarily crisp. Rather, by taking fuzzy membership degrees into account, we are able to quantify to what extent an entity is a member of the fuzzy set representing a property, and how close to it its distractors in their membership of that property. We identified a number of measures of specificity with desirable formal properties (including that they induce a ranking on membership degrees) and addressed the validity of this theoretical framework through an experiment. The results show that the theoretical assumptions are on the right track: Specificity measures are able to predict a significant proportion of the variance in identification time during referential tasks. Furthermore, they are able to account for the odds of selecting the correct referent given a referring expression. From a behavioural perspective, our experiments also confirm that reference resolution by humans is affected by similarity and property membership. 


From the perspective of {\sc reg}, the work presented here serves at least two important purposes: First, it provides a principle revision of the criterion that is typically used by algorithms to determine when a referent has been successfully identified by a description. Second, measures of referential success are also useful to evaluate the output of algorithms. This is especially the case for identification accuracy, where we find a strong relationship between referential success measures and the probability that a referring expression allows a receiver to identify the intended object. 

This work also opens a number of avenues for future work. We are interested in extending our empirical validaiton of specificity measures to more complex scenarios, including descriptions in which a referent is identified by more than one property. Second, we are studying how current {\sc reg} algorithms can be modified to use fuzzy measures of referential success as a stopping criterion, i.e. to determine when a referring expression is likely to provide sufficient information for a user to resolve it successfully.

\section*{Acknowledgments}
This work has been partially supported by the Spanish Government and the European Regional Development Fund - ERDF (Fondo Europeo de Desarrollo Regional - FEDER) under project \mbox{TIN2014-58227-P} \emph{\mbox{Descripci\'on} ling\"u\'istica de informaci\'on visual mediante t\'ecnicas de miner\'ia de datos y computaci\'on flexible}.

\bibliography{specificity}
\bibliographystyle{tex/acl_natbib}
\end{document}